\documentclass[runningheads]{llncs}
\usepackage[T1]{fontenc}
\usepackage{subcaption}
\usepackage{caption}
\usepackage{graphicx}
\usepackage{comment} 
\usepackage{hyperref}
\usepackage{todonotes}
\usepackage{amsmath}
\usepackage{amssymb}
\usepackage{mathtools}
\usepackage{booktabs}	
\usepackage{color, colortbl}
\usepackage{bm}
\usepackage{pifont}%
\newcommand{\cmark}{\ding{51}}%
\newcommand{\xmark}{\ding{55}}%
\usepackage{fancyvrb}
\usepackage{styles/listings-json}
\usepackage{styles/listings-sparql}
\usepackage{styles/listings-turtle}

\usepackage[sort]{cite}
\usepackage{enumitem}

\newlist{questions}{enumerate}{2}
\setlist[questions,1]{label=RQ\arabic*.,ref=RQ\arabic*}
\setlist[questions,2]{label=(\alph*),ref=\thequestionsi(\alph*)}

\definecolor{gray1}{gray}{0.7}

\definecolor{gray2}{gray}{0.85}

\makeatletter
\newcommand{\customlabel}[2]{%
   \protected@write \@auxout {}{\string \newlabel {#1}{{#2}{\thepage}{#2}{#1}{}} }%
   \hypertarget{#1}{#2}
}
\makeatother

\setuptodonotes{inline}
\begin{document}
\title{Lexicalization Is All You Need: Examining the Impact of Lexical Knowledge in a Compositional QALD System}
\titlerunning{Lexicalization Is All You Need}
\author{David Maria Schmidt\inst{1}\orcidID{0000-0001-7728-2884} \and
Mohammad Fazleh Elahi\inst{1}\orcidID{0000-0002-8843-9039} \and
Philipp Cimiano\inst{1}\orcidID{0000-0002-4771-441X}}
\authorrunning{D. M. Schmidt et al.}
\institute{Semantic Computing Group, CITEC, Technical Faculty, Bielefeld University, Bielefeld, Germany\\
\email{\{daschmidt,melahi,cimiano\}@techfak.uni-bielefeld.de}\\
}
\maketitle              %
\begin{abstract}
In this paper, we examine the impact of lexicalization on Question Answering over Linked Data (QALD). It is well known that one of the key challenges in interpreting natural language questions with respect to SPARQL lies in bridging the lexical gap, that is mapping the words in the query to the correct vocabulary elements. We argue in this paper that lexicalization, that is explicit knowledge about the potential interpretations of a word with respect to the given vocabulary, significantly eases the task and increases the performance of QA systems. Towards this goal, we present a compositional QA system that can leverage explicit lexical knowledge in a compositional manner to infer the meaning of a question in terms of a SPARQL query. We show that such a system, given lexical knowledge, has a performance well beyond current QA systems, achieving up to a $35.8\%$ increase in the micro $F_1$ score compared to the best QA system on QALD-9. This shows the importance and potential of including explicit lexical knowledge. In contrast, we show that LLMs have limited abilities to exploit lexical knowledge, with only marginal improvements compared to a version without lexical knowledge. This shows that LLMs have no ability to compositionally interpret a question on the basis of the meaning of its parts, a key feature of compositional approaches. Taken together, our work shows new avenues for QALD research, emphasizing the importance of lexicalization and compositionality. 

\end{abstract}

\section{Introduction}
\label{sec:intro}

Question Answering over Linked Data (QALD)~\cite{shekarpour2016question} is the task of automatically mapping a natural language question to an executable SPARQL query such that relevant information can be retrieved from RDF data sources. 
One of the seven challenges~\cite{unger2014introduction} identified by the authors for the development of QALD systems is handling the lexical gap~\cite{unger2014introduction}, which requires bridging the way users refer to certain natural language terms and the way they are modeled in a given knowledge base. Consider the question \emph{``Who is the mayor of Moscow?''}. In this case, \emph{``mayor''} needs to be interpreted with respect to DBpedia as {\tt dbo:leaderName}\footnote{We use namespace prefixes that are defined as follows: 
\texttt{dbr}: \url{http://dbpedia.org/resource/}, \texttt{dbo}: \url{http://dbpedia.org/ontology/}, \texttt{dbp}: \url{http://dbpedia.org/property/}, 
\texttt{rdfs}: \url{http://www.w3.org/2000/01/rdf-schema\#},
\texttt{rdf}: \url{http://www.w3.org/1999/02/22-rdf-syntax-ns\#}}
to map the question correctly to the following SPARQL query: \texttt{SELECT ?o WHERE \{ dbr:Moscow dbo:leaderName ?o \}}

Another important aspect of QALD is the principle of compositionality. That is, the meaning of a complex expression is determined by the meanings of its parts and the way they are syntactically combined. In the context of QALD, a complex question is represented by a SPARQL query that involves more than one triple pattern, excluding the predicates \emph{rdf:type} or  \emph{rdfs:label}. For example, the SPARQL query of the complex question \emph{``Who is the mayor of the capital of Russia?''} is as follows: \texttt{SELECT ?uri WHERE \{ dbr:Russia dbo:capital ?o . ?o dbo:leaderName ?uri \}}. To handle complex questions, the QALD system requires using compositional reasoning to obtain the answer, which includes multi-hop reasoning, set operations, and other forms of complex reasoning.

Recent approaches based on machine learning models (e.g., deep neural networks~\cite{DBLP:conf/semweb/NikasFT21,hao-etal-2017-end,DBLP:journals/corr/abs-1809-00782,Lukovnikov2017NeuralNQ}, Seq2Seq neural networks~\cite{Omar-etal-2021-universal}, transformers~\cite{DBLP:conf/semweb/LukovnikovF019,neural,liu-etal-2019-generating}, subgraph embeddings \cite{bordes-etal-2014-question}, probabilistic graphical models \cite{DBLP:conf/semweb/HakimovJC17}, bi-direc\-tional LSTMs \cite{DBLP:conf/semco/HakimovJC19}, and tree-LSTMs~\cite{DBLP:journals/corr/abs-2004-13843}) have achieved promising results, and are currently mostly limited to answering simple questions (i.e., only one triple excluding the predicates \emph{rdf:type} and \emph{rdfs:label}). 
To deal with complex queries, Hakimov et al.~\cite{semanticParsing} have proposed an approach that uses \emph{Combinatory Categorial Grammar (CCG)}~\cite{DBLP:books/daglib/0012570} for syntactic representations and typed lambda calculus expressions~\cite{Carpenter1997-CARTS-2} for semantic representations. 
Some approaches~\cite{Unger2011PythiaCM,unger-etal-2010-generating} strongly resemble ours, as the motivation is very similar: using explicit lexical information and \emph{Dependency-based Underspecified Discourse Representation Structures (DUDES)} \cite{dudes1,dudes2} for semantic composition. 
However, these approaches generate all possible combinations of SPARQL queries for a natural language sentence, providing no mechanism for disambiguation; therefore, they produce many logically incorrect SPARQL queries.

Some QALD approaches~\cite{semanticParsing,AskNow2016,DEANNA2012,QAKiS_2012,QAnswer2015} have made only limited use of lexicalization, while others~\cite{Unger2011PythiaCM,EnglishSemantic2021,LexExMachinaQA} have used lexical knowledge but have not systematically investigated its impact. Recently, LLM-based approaches~\cite{LLM_QA_2024,petroni-etal-2019-language,bang-etal-2023-multitask,journals/corr/abs-2306-04181,DBLP:journals/corr/abs-2302-06466,QAKB-2021,gu-su-2022-arcaneqa}  have proven to be powerful tools for NLP tasks. In particular, ChatGPT~\cite{tan2023can,QALD-GPT-3,yih-etal-2016-value} has been shown to be an alternative to traditional QALD approaches. 
To our knowledge, \emph{Generative Pretrained Transformer (GPT)} models have not been tested for their ability to compositionally interpret a question based on the meaning of its parts or the impact of lexical knowledge on their performance. In this paper,
we thus address three research questions and provide the corresponding contributions listed below:
\begin{questions}
\item How can a QA system leverage explicit lexical knowledge? Towards this goal, we present a new compositional QA system that relies on a dependency parse and bottom-up semantic composition. \label{rq:canlex}
\item What is the impact of explicitly given lexical knowledge? Our experimental results show that our compositional system reaches (micro) $F_1$ measures of $0.72$ on the QALD-9 dataset, which outperforms existing state of the art systems on the task by far (+ 35\%). \label{rq:impactlexours}
\item Can Large Language Models also leverage explicit lexical knowledge? Our experiments show that, when encoding lexical knowledge explicitly in the prompt, state-of-the-art LLMs can benefit from such knowledge, improving results. However, they are far from reaching improvements that match the performance of our compositional approach. \label{rq:impactlexllms}
\end{questions}

\section{System Architecture}
\label{sec:systemArch}

\begin{figure}[t]%
\begin{center}
\includegraphics[width=0.95\textwidth]{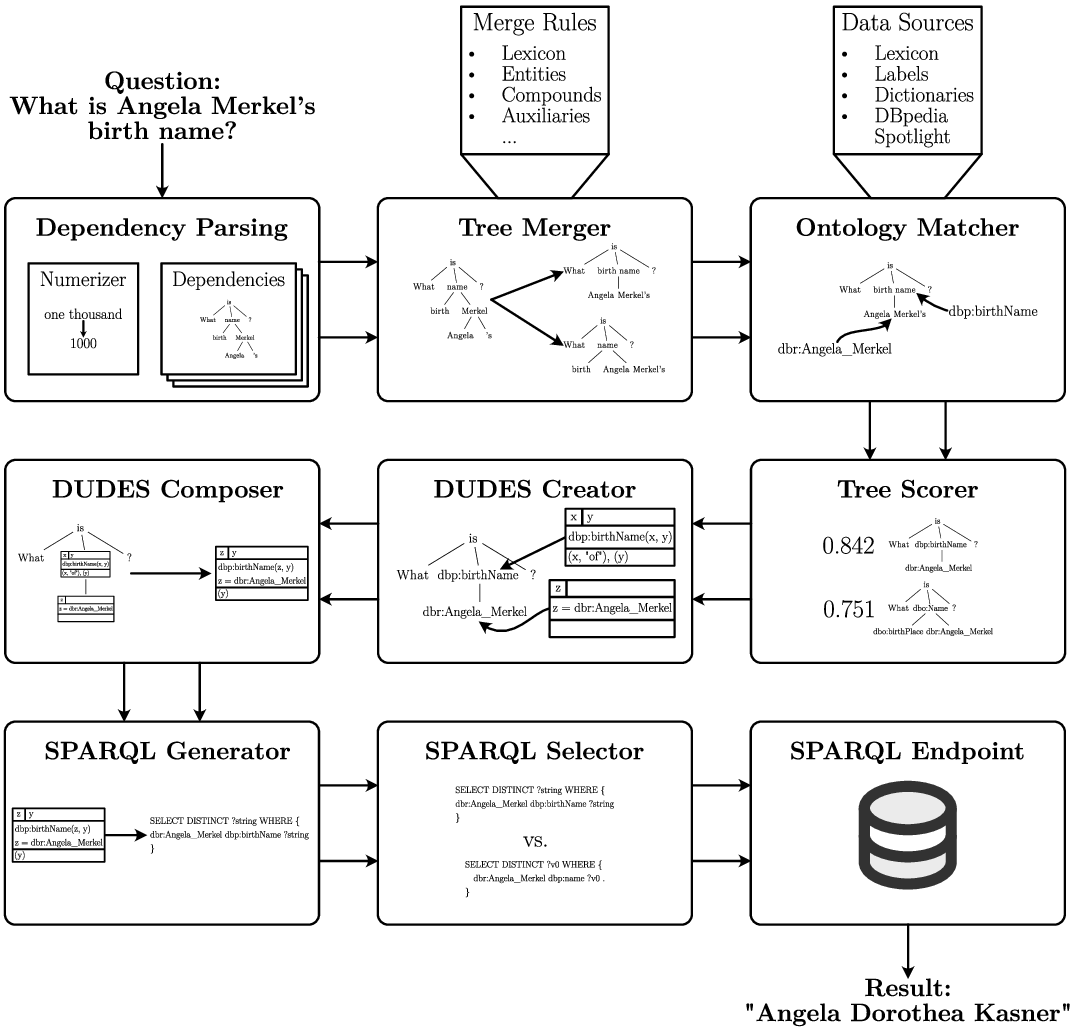}%
\end{center}
\caption{Schema of the compositional question answering approach using DUDES} \label{fig:overview}%
\end{figure}

In this section, we detail a compositional approach to QALD, using \emph{Dependency-based Underspecified Discourse Representation Structures (DUDES)} \cite{dudes1,dudes2} for meaning representation and composition behavior, as well as leveraging explicit lexical knowledge, thus answering \ref{rq:canlex}. The overall architecture of the pipeline is illustrated in Figure \ref{fig:overview} and serves as a blueprint for this section. 
Although we used DBpedia as a reference, our approach can be adapted to any particular ontology and vocabulary by providing a corresponding lexicon.

\subsection{Explicit Lexical Knowledge}
\label{sec:lemon creation}

A necessary prerequisite for our approach is the availability of a Lemon lexicon~\cite{lemon} that describes by which lexical entries the elements (classes, properties) of a particular knowledge base (KB) can be verbalized in a particular language. We rely on the Lemon lexicon format that contains lexical entries and defines how their meaning is captured with respect to a given ontological vocabulary. A \emph{lexical entry} represents a unit of analysis of the lexicon that consists of a set of grammatically related forms and a set of base meanings that are associated with all of these forms.\footnote{Exemplary lexical entries can be found at \url{https://lemon-model.net/}.} The lexicon is context-free in the sense that the possible meanings of words are described independently from their context.

\subsection{Dependency Parsing}
\label{Dependency Parsing}

Our approach relies on a syntactic analysis of an input question by a dependency parser.
To increase the chance that at least one correct dependency tree is generated, which is vital for our approach, we use multiple dependency parsing frameworks (i.e., SpaCy\footnote{\url{https://spacy.io/}} and Stanza/CoreNLP framework\cite{stanza}), configurations, and models\footnote{\texttt{en\_core\_web\_trf} and \texttt{en\_core\_web\_lg}}.
Furthermore, some questions contain textual representations of numbers that need to be translated into their numerical form to be used with, for example, \texttt{FILTER} expressions in SPARQL queries.
However, for entities\footnote{A wide range of subjects, including people, places, organizations, and various concepts, each identified by a unique URI (Uniform Resource Identifier).} that contain textual numbers (e.g., \texttt{dbr:One\_Thousand\_and\_One\_Nights}), this approach might be counterproductive with respect to the entity recognition process.
Therefore, we consider both the converted and the original question in our approach if there is something to be converted. 
To do so, we use the \texttt{numerizer}\footnote{\url{https://github.com/jaidevd/numerizer}} library.

\subsection{Tree Merger}

Matching nodes in the dependency tree to URIs representing entities and properties (a.k.a. \emph{KB Linking}) is a central challenge in our approach. For this purpose, we introduce a number of merging rules over the dependency tree to yield phrases that facilitate matching to KB elements:

\begin{itemize}
  \item \emph{Generic Rules:} Several generic merge rules based on syntactic properties such as dependency tags\footnote{\url{https://universaldependencies.org/u/dep/}} or part-of-speech (POS) tags\footnote{\url{https://universaldependencies.org/u/pos/}} are applied, merging nodes based on, e.g., tags like \texttt{compound} or \texttt{det}, or comparative keywords like \emph{``more''} or \emph{``fewer''}.
  \item \emph{Lexicon Marker Rules:} %
If a lexicon entry matches and includes a marker, the node of that marker (typically an \texttt{ADP} node) is merged into the node that bears the written representation of the corresponding lexical entry. For example, if the lexical entry contains a marker \emph{``of''}, it is merged with the written representation \emph{``birth name''}.%
  \item \emph{Entity Merging Rule:} The presented approach uses several methods for entity recognition (as discussed in Section \ref{sec:ontologymatcher}), and these methods return several candidate entities for a given question. By this rule, the candidate entities, often found at different nodes of the dependency tree, are merged into one node, forming a merged candidate entity. 
As shown in the tree merger step in Figure \ref{fig:overview}, the child node \emph{``Angela''} is merged into its parent node \emph{``Merkel''}, resulting in \emph{``Angela Merkel''}. 
\end{itemize}

\subsection{Ontology Matcher}\label{sec:ontologymatcher}
In this step, entities and properties are assigned to their respective tree nodes where possible. The matching methods are described below.

\paragraph{Property Matching:}
\label{Property Matching}
This matching method focuses on matching the nodes of the tree with DBpedia properties. First, each node of the tree is matched with the written representations of the lexical entries if possible. 
If there is no exact match for a node, the approach tries to find candidate lexical entries by applying several heuristics, such as omitting certain tokens from the node, e.g. by excluding trailing adpositions, which typically do not occur in the canonical forms of lexical entries.
If the marker of a lexical entry matches with a token from a node, the corresponding candidate entry is prioritized.
Finally, if there are remaining ambiguities, the candidate lexical entries are sorted in descending order using a Levenshtein distance-based similarity measure \cite{levenshtein1966binary}.

\paragraph{Entity Matching:}
\label{Entity matching}
To match tree nodes with entities from DBpedia, the approach uses all available \texttt{rdfs:label} information of entities, which are stored in a prefix trie \cite{prefixtrie} for efficient memory representation and lookup of similar labels. The similarities between the entity (e.g., \emph{``Angela Merkel''}) in the tree and the entity labels in DBpedia are calculated using the Levenshtein similarity measure with a threshold set to $0.5$. When both a shorter and a longer text span perfectly match certain labels, the longer match is generally prioritized higher. 
As an off-the-shelf solution, we also include the entity recognition results of DBpedia Spotlight \cite{dbpedia-spotlight} into the set of considered candidates to increase the chance of a correct match. 

\subsection{Tree Scorer}
Each node of the tree is assigned a score based on matched properties or entities. Additionally, it considers the total number of tree nodes, as well as special terms and keywords which are neither properties nor entities. Two types of matching are taken into account: (i) exact matches, 
and (ii) matches under relaxed conditions. %
The scoring relies on a weighted average of three different scores: (i) fraction of nodes with exact matches (weight $3$), (ii) fraction of nodes with matches under relaxed conditions (weight $1$), and (iii) ratio of the number of nodes to the number of nodes in the dependency tree before merging nodes (weight $2$).
For (i) and (ii), single node weights (i.e., the number of tokens a node comprises) are multiplied with different multipliers based on whether the node has a matching lexical entry (multiplier $1.0$), a matching entity or is a numeral ($0.9$) or is a special word like an \texttt{ASK} keyword, a comparative or \emph{``in''} ($0.8$). Then, the weight is multiplied with that multiplier and added to a total sum. In the end, this sum is compared to the sum of all weights, forming the score value. 
The weighted average of these three scores forms the total score of a tree, according to which the trees are then prioritized in processing.

\subsection{DUDES Creator}\label{sec:dudescreator}

Now that we have a tree with KB elements (e.g., entities and properties) assigned to the nodes, the next task is to create \emph{Dependency-based Underspecified Discourse Representation Structures (DUDES)} \cite{dudes1,dudes2}, which are used to compose the atomic meanings of the tree nodes.
Our approach is slightly different from the latest version of DUDES \cite{dudes2} as we modify it for use with dependency trees instead of \emph{Lexicalized Tree Adjoining Grammar (LTAG)} trees \cite{Joshi1997,schabes-joshi-1988-earley} and do not make use of subordination relations yet:

\begin{definition}[Dependency-based Underspeciﬁed Discourse Representation Structures (DUDES) \cite{dudes2}]
    A DUDES is a triple $(v, D, S)$ where:
    \begin{itemize}
        \item $v \in U \cup \{\epsilon\}$ is the main variable (also called referent marker or distinguished variable) where $\epsilon$ represents the absence of a main variable
        \item $D = (U, C)$ is a Discourse Representation Structure (DRS) \cite{dudes2,drs1,drs2} with
        \begin{itemize}
            \item set of variables $U$ (also called discourse universe or referent markers)
            \item set of conditions $C$ over variables $U$
        \end{itemize}
        \item $S$ is a set of selection pairs of the form $(v \in U, m)$ with $v$ being a variable from $U$ and $m$ being a marker word for that variable with $\epsilon$ representing the empty marker, i.e. no marker being connected to that variable. Instead of writing $\epsilon$, the second tuple component can also just be left out.
    \end{itemize}
\end{definition}

\begin{figure}%
    \centering%
    \begin{subfigure}[t]{0.3\textwidth}
    \centering
    \includegraphics[width=\textwidth]{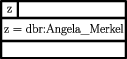}
    \caption{Entity DUDES \texttt{dbr:Angela\_Merkel} with main variable $z$, condition $z = \mathit{dbr{:}Angela\_Merkel}$ and no selection pairs} \label{fig:dudesent}
    \end{subfigure}
    \begin{subfigure}[t]{0.3\textwidth}
    \centering
    \includegraphics[width=0.89\textwidth]{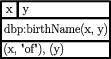}
    \caption{Property DUDES \texttt{dbo:birthName} with main variable $x$, condition $\mathit{dbo{:}birthName}(x,y)$ and selection pairs $(x, ``of")$ and $(y, \epsilon)$} \label{fig:dudesprop}
    \end{subfigure}
    \begin{subfigure}[t]{0.3\textwidth}
    \centering
    \includegraphics[width=0.79\textwidth]{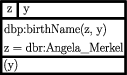}
    \caption{Composition of Figure \ref{fig:dudesent} and \ref{fig:dudesprop} using selection pair $(x, ``of")$. Results in condition $\mathit{dbo{:}birthName}(z,y) \land z = \mathit{dbr{:}Angela\_Merkel}$} \label{fig:dudescomb}
    \label{fig:dudescomp}
    \end{subfigure}
\caption{Illustration of exemplary DUDES and their composition} \label{fig:dudesexamples} %
\end{figure}

\paragraph{Entity DUDES:} 
The simplest case of representing KB elements from the tree as a DUDES is representing entities (i.e., \emph{Entity DUDES}). 
In Entity DUDES, an entity is assigned to a variable, for example, by adding a simple expression such as $z = \mathit{dbr{:}Angela\_Merkel}$. A full example for entity \texttt{dbr:Angela\_Merkel} is illustrated in Figure \ref{fig:dudesent}.

\paragraph{Property DUDES:} 
In contrast to entities, properties have variables that are intended to be replaced by entities or combined with other properties during DUDES composition.
Additionally, variables can be restricted to certain markers that correspond to the subject or object position of the property. An example of the property \texttt{dbo:birthName} is shown in Figure \ref{fig:dudesprop}. Note the variable $x$ is associated with the marker \emph{``of''} as another way of disambiguation which is however not used in this example. For, e.g., \textit{``What is the birth name of Angela Merkel?''}, it would instead be used to determine the subject of the property.

\subsection{DUDES Composer}

The composition operation of DUDES \cite{dudes2} can be defined as follows:

\begin{definition}[DUDES Composition]
    Let $d_1 = (v_1, D_1 = (U_1, C_1), S_1)$, $d_2 = (v_2, D_2 = (U_2, C_2), S_2)$ be two DUDES with disjoint variable sets, i.e. $U_1 \cap U_2 = \emptyset$. The DUDES composition operation $\odot$ for substituting $d_1$ into $d_2$ using selection pair $p = (x \in U_2, m) \in S_2$ and resulting in a composed DUDES ${d_c = (v_c, D_c = (U_c, C_c), S_c)}$, written $d_c = d_1 \overset{p}{\odot} d_2$, is defined as follows:
    \begin{multicols}{3}
    \noindent
        \begin{align*}
            &U_c = U_2[x \coloneq v_1] \cup U_1 \\
            &C_c = C_2[x \coloneq v_1] \cup C_1 
        \end{align*}
        \begin{align*}
            &S_c = (S_2 \cup S_1) \setminus p 
        \end{align*}
        \begin{align*}
           v_c = \begin{cases}
                v_1 & \text{if } x = v_2 \\
                v_2 & \text{else}
            \end{cases}
        \end{align*}
    \end{multicols}
\end{definition}
An example composition of the two DUDES (i.e., Figure \ref{fig:dudesent} and \ref{fig:dudesprop}) is shown in Figure  \ref{fig:dudescomp}.
We apply a bottom-up composition strategy, merging child nodes into their parent nodes. DUDES compositions are performed until there is only a single composed DUDES (i.e., final DUDES) left at the root of the tree, representing the meaning of the whole question.
For choosing a selection pair for composition, different heuristics and data sources are used. %
For example, in the case of modifier nodes, the parent DUDES is merged into the child DUDES and not the other way around.  
Additionally, the syntactic frames of the lexical entries, POS and dependency tags are used for selection pair determination.
Instead of calculating all combinations at once, our approach composes one final DUDES at a time, limiting the effect of combinatorial explosion that would otherwise substantially increase the memory footprint.

\subsection{SPARQL Generator}
\label{SPARQL Generator}
The logical expressions of the final DUDES represent the meaning of the given natural language question. Therefore, they are used to create a corresponding SPARQL query. 
For instance, the DUDES in Figure  \ref{fig:dudescomb} shows the triple pattern (i.e., \texttt{dbr:Angela\_Merkel dbo:birthName ?y}) for the question \emph{``What is Angela Merkel's birth name?''}. %
The triple patterns of DUDES contain entities (or literals), and variables. Our approach uses the Z3 SMT solver \cite{z3} to determine which variables in the final DUDES are bound to some values. 
Due to our combinatorial approach to dealing with ambiguities, multiple SPARQL queries are typically generated, from which one is selected by the  SPARQL selector.

\subsection{SPARQL Selector}
\label{sec:SPARQL Selector}
For selecting the best SPARQL query, we use an LLM-based approach \cite{GenerativeRelation-2021} trained to compare two queries, using the encoder of \texttt{flan-t5-small} \cite{flant5} as a base model. %
As single LLM-based comparison results are still unreliable, various aggregation strategies are evaluated which make a final decision based on the pairwise comparisons of all candidate queries.

For each of the two candidate queries of a comparison, the model is given the input question, the candidate query, the number of its results, and the final DUDES. From this information, two output features are generated, representing the confidence in the respective queries. 
In order to reflect in the output how much better one query is than another, the model is trained to predict the $F_1$ scores of the respective queries.\footnote{To avoid high numbers of false positives affecting the micro $F_1$ score, queries with a false to true positive ratio of $10:1$ or worse are clamped to $0.0$ in the training data.}
We evaluate different strategies and configurations to select the final query:

\begin{itemize}
    \item \textbf{BestScore:} %
    Theoretically possible performance of our approach, selecting best queries based on their true $F_1$ score, clamped like in training data.
    
    \item $\textbf{MostWins}_{p\%}^{\mathit{top}~n}$\textbf{:} 
    Compares candidate queries pairwise per question and selects the query that ``wins'' the most comparisons (with margin of $>p\%$).
    \item $\textbf{Accum}_{\mathit{logits}/\mathit{sigmoid}}^{\mathit{top}~n}$\textbf{:} For each candidate query, the model outputs are accumulated, either \emph{logits} or \emph{sigmoid} values, largest value is chosen.
\end{itemize}

If an exponent $\mathit{top}~n$ is given, the top $n$ models (based on training micro $F_1$ score) are evaluated with their outputs summed together. Otherwise, single models are evaluated and presented with mean and standard deviation. Additionally, queries with no results or with too many results (threshold is the largest number of results for a train question + $10\%$) are discarded.

\section{Experimental Setup}
\label{sec:exp}
The experiments were conducted using the well-known QALD-9 benchmark~\cite{Usbeck2023}. The dataset contains questions in multiple languages, along with corresponding SPARQL queries and answers from DBpedia. 
For the experiment, we followed Unger et al. \cite{dblexipedia} 
to manually create a lexicon covering the vocabulary elements in the training and test section of QALD-9. In particular, we created a total of {$599$} lexical entries for five syntactic frames~\cite{DBLP:journals/ws/CimianoBMS11}: {$311$} lexical entries for  \texttt{NounPPFrame}, {$96$} lexical entries for \texttt{TransitiveFrame},  {$143$}  lexical entries for  \texttt{InTransitivePPFrame}, {$28$} lexical entries for \texttt{AdjectivePredicateFrame}, and {$21$} lexical entries for \texttt{AdjectiveSuperlativeFrame}. The time required for creating a lexical entry was approximately $2$--$5$ minutes depending on the syntactic frame. The total time required to create our lexicon was approximately $16$ hours. 
All experiments\footnote{Software artifact: \url{https://doi.org/10.5281/zenodo.12610054}, System: AMD Ryzen 9 7900X3D, 96GB RAM, NVIDIA GeForce RTX 4070, Arch Linux 6.9.2-arch1-1, Python 3.12.3, $12$ parallel processes, total (elapsed real time) timeout of $10800$s for test benchmark, DBpedia version: 2016-10 \url{https://downloads.dbpedia.org/2016-10/core-i18n/en/}} were conducted for the English part of QALD-9 only.

\subsection{SPARQL Selection Model Training}
In Section \ref{SPARQL Generator}, we presented an LLM-based SPARQL selection approach for disambiguation of the generated SPARQL queries. In order to have training, validation and test data for the query selection model, we ran our approach for about 24h on the QALD-9 benchmark and saved all generated candidate queries, separated by train and test questions. 
Afterwards, the training data was randomly split into $90\%$ training questions and $10\%$ validation questions.

The corresponding final data elements were generated in three steps for each part of the data (i.e., training, validation, and test). Each step involved generating $100$ comparisons for each question and used for training in a symmetric way to avoid some general preference of the model for the first or second query. %
Comparisons between queries with (i) an $F_1$ score $\geq 0.01$ and (ii) an $F_1$ score $< 0.01$ were added to the training data. Additionally, mixed comparisons with one query with an $F_1$ score $\geq 0.01$ and one with an $F_1$ score $< 0.01$ were added.

In order to fine-tune the \texttt{google/flan-t5-small} model from the Huggingface transformers library \cite{huggingface}, we first ran a hyperparameter optimization with $34$ trials using Optuna \cite{optuna}, with an epoch search space between $1$ and $5$, an initial learning rate between $1e^{-5}$ and $1e^{-4}$ (logarithmic scale) and using a lambda learning rate scheduler with a lambda between $0.9$ and $1.0$ (logarithmic scale). As an optimizer, we used Adam \cite{adam} and trained $10$ models using the parameters of the trial with the lowest validation loss discovered during the hyperparameter optimization. Each training was performed on a single Nvidia A40 with a batch size of $64$. These $10$ models were then used for evaluation.

\subsection{Experiments with GPT}
\label{sec:expgpt}

We compared different GPT~\cite{ouyang2022training} models to our compositional approach. 
The previous research on QALD with GPT-3 \cite{QALD-GPT-3} evaluated using the QALD-9 test dataset in three modes: zero-shot, few-shot, and fine-tuned model. In the zero-shot scenario, GPT-3 generated many invalid queries. Performance increased with the five-shot approach and even more with fine-tuning.

All of our experiments with GPT models are performed separately with $5$ different prompts describing the task. The first prompt below
has been hand-crafted. Afterwards, four additional prompts have been generated using ChatGPT using the prompt \emph{You are a world-class prompt engineer. Refine this prompt: <initial prompt>}. The resulting prompts used in our experiments are therefore:

\begin{enumerate}[before=\normalfont\scriptsize]
    \item You are a system which creates SPARQL queries for DBPEDIA from 2016-10 from natural language user questions. You answer just with SPARQL queries and nothing else.
    \item Generate SPARQL queries from user questions for DBpedia from October 2016. Answer solely with SPARQL queries.
    \item Develop a system capable of generating SPARQL queries for DBPedia based on user questions in natural language, with a knowledge base updated until October 2016. The system should exclusively respond with SPARQL queries and no additional information.
    \item Craft SPARQL queries from October 2016 based on user questions in natural language, exclusively dedicated to extracting information from DBpedia. Your responses should consist solely of SPARQL queries.
    \item Create SPARQL queries to generate responses to user questions by interpreting natural language queries, specifically targeting DBpedia, beginning from October 2016.
\end{enumerate}

For our experiments we, prompted GPT-3.5-Turbo and GPT-4 in a zero-shot fashion and evaluated them on the entire training and test datasets of QALD-9. For the fine-tuned GPT-3.5-Turbo-0125 models, 
$10\%$ of the original training dataset has been excluded and used as a validation dataset for fine-tuning. 
All experiments are executed with temperature $0$ as well as both with and without lexical information in the prompt, i.e. lexical information was also present during fine-tuning.  To the best of our knowledge, no work has investigated the impact of using lexical information (which is crucial for state-of-art performance for QALD) on the benchmark performance in this way yet. 
For the experiments with lexical information (as detailed in \ref{sec:lemon creation}), we shorten the structure of lexical entries (the structure is detailed in previous work~\cite{DBLP:conf/iswc/Benz2020}) for prompting and training, as the lexical entries are not well-suited for direct usage.\footnote{Datasets generated this way together with the shortened lexical entries can be found in our software artifact: \url{https://doi.org/10.5281/zenodo.12610054}} This shortened representation consists primarily of pairs of field names and their values, e.g. \emph{``Canonical form: birth name''} or \emph{``Reference: dbp:birthName''}. To fit into the context window of the used models, we restrict the entries appended to the prompt to entries which are relevant to the question.\footnote{This means where written representations occur in the question and ontology URIs in the gold standard query.} Therefore, only the ability of GPT models to put the pieces together is tested, not whether they select the right entry from a much larger lexicon. The comparison is therefore not fair as our approach figures out the relevant lexical entries itself. The numbers presented in the evaluation section are therefore to be interpreted as ``upper bounds'' on the performance. 
The OpenAI API has been used for fine-tuning, as the model is not publicly available. The configurable hyperparameters \textit{batch-size}, \textit{learning rate multiplier} as well as \textit{number of epochs} were optimized using the \textit{auto} setting.%

\section{Evaluation}
\label{sec:eval}
The evaluation of our approach is presented in three categories: \emph{Single Model}, \emph{Multi Model}, and \emph{Upper Bounds}. The first category shows mean and standard deviation across the $10$ trained models for SPARQL selection strategies using a single model. As these results show a high standard deviation for micro scores, we also evaluated the effect of bundling the outputs of multiple models.
For bundling the model outputs (i.e., $\mathit{top}~n$ models), we focused on the strategies and models performing best on the training dataset of QALD-9. 
We also included \emph{BestScore} strategies in the \emph{Upper Bounds} category, demonstrating the highest achievable scores (i.e., upper bounds on the query selection model performances). Good scores in the \emph{BestScore}/\emph{Upper Bounds} category therefore indicate that the pipeline in principle generates the correct results, but those queries are not always identified during query selection. However, selecting the best query can be considered a much easier task than generating it from scratch, rendering these scores still reasonably realistic. Nevertheless, when comparing to other approaches, only the best performances achieved by a regular query selection strategy are used for fairness reasons. %

\begin{table}[ht]
    \vspace{-5mm}
    \caption{Results for English QALD-9 test dataset after $3$ hours of elapsed real time. P refers to Precision, R to Recall, and $F_1$ to $F_1$ score. The best results of each category are marked in bold. \emph{``(single)''} means running the benchmark without evaluating LLM strategy performance at the same time, reducing the corresponding overhead.}
    \small
    \centering    
    \resizebox{\textwidth}{!}{%
    \begin{tabular}{l||c|c|c||c|c|c}
    \toprule
    & \multicolumn{3}{c|}{Micro} & \multicolumn{3}{c}{Macro} \\
    \multicolumn{1}{c||}{Strategy} & \multicolumn{1}{c}{$F_1\pm\sigma$} & \multicolumn{1}{c}{P $\pm\sigma$} & \multicolumn{1}{c|}{R $\pm\sigma$} & \multicolumn{1}{c}{$F_1\pm\sigma$} & \multicolumn{1}{c}{P $\pm\sigma$} & \multicolumn{1}{c}{R $\pm\sigma$} \\
    \midrule
    \multicolumn{7}{c}{Single Model} \\\midrule
    $\text{Accum}_{logits}$ & $0.30\pm 0.10$ & $0.24\pm 0.11$ & $0.59\pm 0.25$ & $0.31\pm 0.02$ & $0.31\pm 0.02$ & $0.34\pm 0.02$ \\
    $\text{Accum}_{sigmoid}$ & $0.39\pm 0.13$ & $0.31\pm 0.11$ & $0.61\pm 0.20$ & $\bm{0.32\pm 0.01}$ & $\bm{0.32\pm 0.01}$ & $\bm{0.35\pm 0.02}$ \\
    $\text{MostWins}_{0.0}$ & $0.29\pm 0.08$ & $0.19\pm 0.06$ & $\bm{0.65\pm 0.10}$ & $0.30\pm 0.02$ & $0.30\pm 0.01$ & $0.34\pm 0.01$ \\
    $\text{MostWins}_{0.1}$ & $0.33\pm 0.08$ & $0.23\pm 0.07$ & $0.62\pm 0.14$ & $0.31\pm 0.01$ & $0.31\pm 0.01$ & $0.34\pm 0.01$ \\
    $\text{MostWins}_{0.25}$ & $0.40\pm 0.16$ & $0.32\pm 0.22$ & $0.65\pm 0.11$ & $\bm{0.32\pm 0.01}$ & $\bm{0.32\pm 0.01}$ & $\bm{0.35\pm 0.02}$ \\
    $\text{MostWins}_{0.5}$ & $0.36\pm 0.13$ & $0.28\pm 0.10$ & $0.58\pm 0.21$ & $0.31\pm 0.02$ & $\bm{0.32\pm 0.02}$ & $0.34\pm 0.02$ \\
    $\text{MostWins}_{0.75}$ & $\bm{0.43\pm 0.19}$ & $\bm{0.38\pm 0.23}$ & $0.56\pm 0.21$ & $\bm{0.32\pm 0.02}$ & $\bm{0.32\pm 0.02}$ & $0.34\pm 0.02$ \\
    $\text{MostWins}_{0.9}$ & $0.42\pm 0.18$ & $0.36\pm 0.22$ & $0.56\pm 0.21$ & $\bm{0.32\pm 0.02}$ & $\bm{0.32\pm 0.02}$ & $0.34\pm 0.02$ \\ \midrule
    
    \multicolumn{7}{c}{Multi Model} \\\midrule
    $\text{MostWins}_{0.75}^{\mathit{top}~2}$ & \multicolumn{1}{c|}{$\bm{0.72}$} & \multicolumn{1}{c|}{$\bm{0.77}$} & \multicolumn{1}{c||}{$0.67$} & \multicolumn{1}{c|}{$0.32$} & \multicolumn{1}{c|}{$0.32$} & \multicolumn{1}{c}{$0.33$} \\ 
    $\text{MostWins}_{0.9}^{\mathit{top}~2}$ & \multicolumn{1}{c|}{$0.64$} & \multicolumn{1}{c|}{$0.61$} & \multicolumn{1}{c||}{$0.67$} & \multicolumn{1}{c|}{$0.32$} & \multicolumn{1}{c|}{$0.32$} & \multicolumn{1}{c}{$0.34$} \\ 

    $\text{MostWins}_{0.75}^{\mathit{top}~3}$ & \multicolumn{1}{c|}{$\bm{0.72}$} & \multicolumn{1}{c|}{$\bm{0.77}$} & \multicolumn{1}{c||}{$0.67$} & \multicolumn{1}{c|}{$0.32$} & \multicolumn{1}{c|}{$\bm{0.33}$} & \multicolumn{1}{c}{$0.33$} \\ 
    $\text{MostWins}_{0.9}^{\mathit{top}~3}$ & \multicolumn{1}{c|}{$0.65$} & \multicolumn{1}{c|}{$0.64$} & \multicolumn{1}{c||}{$\bm{0.68}$} & \multicolumn{1}{c|}{$0.32$} & \multicolumn{1}{c|}{$\bm{0.33}$} & \multicolumn{1}{c}{$\bm{0.35}$} \\ 

    $\text{MostWins}_{0.75}^{\mathit{top}~5}$ & \multicolumn{1}{c|}{$0.59$} & \multicolumn{1}{c|}{$0.52$} & \multicolumn{1}{c||}{$\bm{0.68}$} & \multicolumn{1}{c|}{$\bm{0.33}$} & \multicolumn{1}{c|}{$\bm{0.33}$} & \multicolumn{1}{c}{$\bm{0.35}$} \\ 
    $\text{MostWins}_{0.9}^{\mathit{top}~5}$ & \multicolumn{1}{c|}{$0.59$} & \multicolumn{1}{c|}{$0.53$} & \multicolumn{1}{c||}{$\bm{0.68}$} & \multicolumn{1}{c|}{$\bm{0.33}$} & \multicolumn{1}{c|}{$\bm{0.33}$} & \multicolumn{1}{c}{$\bm{0.35}$} \\ 

    $\text{MostWins}_{0.75}^{\mathit{top}~10}$ & \multicolumn{1}{c|}{$0.59$} & \multicolumn{1}{c|}{$0.53$} & \multicolumn{1}{c||}{$0.67$} & \multicolumn{1}{c|}{$0.32$} & \multicolumn{1}{c|}{$\bm{0.33}$} & \multicolumn{1}{c}{$0.34$} \\ 
    $\text{MostWins}_{0.9}^{\mathit{top}~10}$ & \multicolumn{1}{c|}{$0.62$} & \multicolumn{1}{c|}{$0.57$} & \multicolumn{1}{c||}{$\bm{0.68}$} & \multicolumn{1}{c|}{$\bm{0.33}$} & \multicolumn{1}{c|}{$\bm{0.33}$} & \multicolumn{1}{c}{$0.34$} \\ 
    \midrule
    
    \multicolumn{7}{c}{Upper Bounds} \\\midrule
    BestScore & \multicolumn{1}{c|}{$0.81$} & \multicolumn{1}{c|}{$\bm{0.98}$} & \multicolumn{1}{c||}{$\bm{0.69}$} & \multicolumn{1}{c|}{$0.37$} & \multicolumn{1}{c|}{$0.38$} & \multicolumn{1}{c}{$0.38$} \\
    BestScore (single) & \multicolumn{1}{c|}{$\bm{0.85}$} & \multicolumn{1}{c|}{$0.95$} & \multicolumn{1}{c||}{$0.76$} & \multicolumn{1}{c|}{$\bm{0.51}$} & \multicolumn{1}{c|}{$\bm{0.51}$} & \multicolumn{1}{c}{$\bm{0.54}$} \\
    \bottomrule
    \end{tabular}}%
    \label{tab:qaldresults}%
    \vspace{-5mm}
\end{table}%
During evaluation, all strategies with all 10 SPARQL selection models were tested simultaneously to ensure they were evaluated on the same generated queries. However, as we limited the elapsed real time to $3$ hours, this imposed a high overhead w.r.t. single strategies. Evaluating just one strategy and model at once would likely have achieved better results due to more tested candidates.\footnote{Generated candidate queries per question: up to $5439$, mean: $87.87 \pm 495.02$.} Illustrating the theoretical potential of the generated queries, a second evaluation with just \emph{BestScore} being executed for $3$ hours was conducted (marked with \emph{``(single)''} in Table \ref{tab:qaldresults}), generating $815473$ instead of $10552$ queries and increasing scores from $0.37$ to $0.51$ (macro $F_1$).\footnote{Generated candidate queries per question: up to $59310$, mean: $5660.98 \pm 8268.85$.} %

Table \ref{tab:qaldresults} shows that the multi-model strategies generally outperform single-model strategies, e.g., $0.43$ vs. $0.72$ vs. $0.85$ for the micro $F_1$ scores of single-model, multi-model, and upper bound strategies, respectively.
More precisely, single-model strategies on average achieve only about half of the upper bound performances, although they exhibit a high standard deviation, indicating their potential to yield substantially different results based on the specific trained model chosen for evaluation.
However, aggregating the outputs of multiple models to select a query appears to combine the strengths of the bundled models without being affected by their weaknesses. This results in comparably stable performance across different numbers of bundled models (e.g., $0.59$ to $0.72$ for micro $F_1$ scores).
In contrast, the macro $F_1$ scores are consistent across both single and multi-model strategies, indicating strong overall performance.%

\begin{table}[ht]
    \vspace{-5mm}
    \caption{GPT results for QALD-9 test dataset. P refers to Precision, R to Recall, $F_1$ to $F_1$ score, FT to fine-tuned, and Pr\# to the prompt number. The best results of each experiment are marked in bold, total best scores of a category are underlined.}
    \small
    \centering
    \resizebox{\textwidth}{!}{%
    \begin{tabular}{l|cc||ccc|ccc||ccc|ccc}
        \toprule
        & & & \multicolumn{6}{c||}{Without Lexicon} & \multicolumn{6}{c}{With Lexicon} \\
        & & & \multicolumn{3}{c|}{Micro} & \multicolumn{3}{c||}{Macro} & \multicolumn{3}{c|}{Micro} & \multicolumn{3}{c}{Macro} \\
        Model & FT & Pr\# & $F_1$ & P & R & $F_1$ & P & R & $F_1$ & P & R & $F_1$ & P & R \\ \midrule
        GPT-3.5-Turbo & \xmark & 1 & $\bm{0.15}$ & $0.19$ & $\bm{0.12}$ & $0.10$ & $0.11$ & $0.12$ & $0.13$ & $0.09$ & $\bm{0.21}$ & $0.25$ & $\bm{0.27}$ & $0.27$ \\
        GPT-3.5-Turbo & \xmark & 2 & $\bm{0.15}$ & $\bm{0.21}$ & $\bm{0.12}$ & $0.11$ & $0.10$ & $0.12$ & $0.13$ & $0.09$ & $0.20$ & $\bm{0.26}$ & $0.26$ & $\bm{0.31}$ \\
        GPT-3.5-Turbo & \xmark & 3 & $0.12$ & $0.18$ & $0.10$ & $\bm{0.13}$ & $\bm{0.13}$ & $\bm{0.15}$ & $0.06$ & $0.04$ & $0.13$ & $0.21$ & $0.21$ & $0.25$ \\
        GPT-3.5-Turbo & \xmark & 4 & $0.08$ & $0.06$ & $\bm{0.12}$ & $0.12$ & $0.11$ & $\bm{0.15}$ & $0.04$ & $0.02$ & $0.15$ & $0.23$ & $0.22$ & $0.27$ \\
        GPT-3.5-Turbo & \xmark & 5 & $0.10$ & $0.17$ & $0.07$ & $0.05$ & $0.05$ & $0.05$ & $\bm{0.16}$ & $\bm{0.45}$ & $0.10$ & $0.12$ & $0.12$ & $0.14$ \\ \midrule
        GPT-4 & \xmark & 1 & $0.22$ & $0.29$ & $0.18$ & $0.26$ & $0.27$ & $0.28$ & $\underline{\bm{0.35}}$ & $0.81$ & $0.22$ & $\bm{0.40}$ & $\bm{0.40}$ & $\bm{0.42}$ \\
        GPT-4 & \xmark & 2 & $\bm{0.34}$ & $\bm{0.68}$ & $\underline{\bm{0.23}}$ & $\bm{0.28}$ & $\bm{0.29}$ & $\bm{0.30}$ & $0.31$ & $\bm{0.87}$ & $0.19$ & $0.39$ & $\bm{0.40}$ & $0.40$ \\
        GPT-4 & \xmark & 3 & $0.22$ & $0.25$ & $0.20$ & $0.26$ & $0.27$ & $0.28$ & $0.12$ & $0.09$ & $\bm{0.21}$ & $0.39$ & $\bm{0.40}$ & $\bm{0.42}$ \\
        GPT-4 & \xmark & 4 & $0.32$ & $0.65$ & $0.21$ & $0.25$ & $0.26$ & $0.29$ & $0.12$ & $0.08$ & $0.20$ & $0.38$ & $0.38$ & $0.40$ \\
        GPT-4 & \xmark & 5 & $0.19$ & $0.17$ & $0.21$ & $0.20$ & $0.19$ & $0.25$ & $0.24$ & $0.32$ & $0.20$ & $0.26$ & $0.26$ & $0.29$ \\ \midrule
        GPT-3.5-Turbo & \cmark & 1 & $0.28$ & $0.81$ & $0.17$ & $0.24$ & $0.24$ & $0.25$ & $0.22$ & $0.28$ & $0.18$ & $\underline{\bm{0.42}}$ & $\underline{\bm{0.43}}$ & $0.42$ \\
        GPT-3.5-Turbo & \cmark & 2 & $\underline{\bm{0.35}}$ & $0.88$ & $\bm{0.22}$ & $0.24$ & $0.25$ & $0.25$ & $0.11$ & $0.08$ & $0.16$ & $0.40$ & $0.42$ & $0.42$ \\
        GPT-3.5-Turbo & \cmark & 3 & $0.26$ & $0.50$ & $0.18$ & $\bm{0.25}$ & $\bm{0.26}$ & $\bm{0.26}$ & $0.09$ & $0.06$ & $0.15$ & $0.37$ & $0.37$ & $0.40$ \\
        GPT-3.5-Turbo & \cmark & 4 & $\underline{\bm{0.35}}$ & $0.91$ & $\bm{0.22}$ & $0.23$ & $0.24$ & $0.24$ & $0.10$ & $0.07$ & $\bm{0.21}$ & $0.41$ & $0.41$ & $\underline{\bm{0.44}}$ \\
        GPT-3.5-Turbo & \cmark & 5 & $0.34$ & $\underline{\bm{0.92}}$ & $0.21$ & $0.24$ & $0.25$ & $0.25$ & $\bm{0.31}$ & $\bm{0.87}$ & $0.19$ & $0.38$ & $0.38$ & $0.40$ \\ \bottomrule
    \end{tabular}}
    \label{tab:gptqaldresults}
    \vspace{-5mm}
\end{table} 
\paragraph{Comparison with GPT models:}
The results of our experiments (as shown in Table \ref{tab:gptqaldresults}) with QALD-9 and GPT models \cite{ouyang2022training} 
are examined with two different objectives: comparing the GPT performance with our approach, i.e. Table \ref{tab:qaldresults}, (research question \ref{rq:impactlexours}) and examining the effect of providing the lexical entries in the prompt (research question \ref{rq:impactlexllms}).
Table \ref{tab:gptqaldresults} shows the $F_1$ scores of GPT-3.5-Turbo models with and without fine-tuning, as well as GPT-4 without fine-tuning, with and without a lexicon. %
Regarding the first objective (\ref{rq:impactlexours}), 
our approach outperforms GPT models in terms of micro $F_1$ score, achieving $0.72$ compared to $0.35$ of the best-performing GPT model.%

In contrast, the total best macro $F_1$ score of all evaluated GPT models outperforms the macro $F_1$ scores of our approach ($0.33$ vs. $0.42$). However, this is only true for models that were provided with the correct lexical entries in the prompt, a substantial simplification compared to our approach which has to determine the relevant lexical entries from the whole lexicon. Without this advantage, all evaluated GPT models are outperformed by our approach, as the macro $F_1$ score does not exceed $0.28$ then. Additionally, our upper bounds show scores up to $0.51$. %
However, our current query selection models do not reach these scores, which remains to be solved in future work.

Regarding the second objective (\ref{rq:impactlexllms}), adding lexical entries to the prompt improves the macro scores in almost all cases, whereas the effect on the micro scores is mixed. For the non-fine-tuned GPT-3.5-Turbo model, the macro $F_1$ scores even doubles from $0.13$ to $0.26$. This effect is similarly large for GPT-4 ($0.28$ vs. $0.40$) and fine-tuned GPT-3.5-Turbo ($0.25$ vs. $0.42$). 

\begin{table}[ht]
    \vspace{-5mm}
    \caption{Comparison with SOTA evaluated on the QALD-9 test dataset.}
    \small
    \centering
    \begin{tabular}{|l|c|c|c|}\hline
    	 \textbf{QALD System}   &\textbf{Micro Precision} & \textbf{Micro Recall} &  \textbf{Micro $\bm{F_1}$ Score} \\ 
    		\hline    
       Galactica~\cite{glaese2022improving} &{$0.14$} & {$0.02$}  & {$0.03$} \\
       Elon~\cite{qald9} &{$0.04$} & {$0.05$}  & {$0.10$} \\
       QASystem~\cite{qald9} &{$0.09$} & {$0.11$}  & {$0.20$} \\
       Falcon 1.0~\cite{sakor-etal-2019-old}   &{$0.23$} & {$0.23$}  & {$0.23$} \\
       WDAqua-core1~\cite{DBLP:journals/semweb/DiefenbachBSM20} &{$0.26$} & {$0.26$}  & {$0.28$} \\
       EDGQA~\cite{Xixin-etal-2021-EDG-Based} &{$0.31$} & {$0.40$}  & {$0.32$} \\
       TeBaQA~\cite{Vollmers-etal-2021-Knowledge} &{$0.24$} & {$0.24$}  & {$0.37$} \\
       gAnswer~\cite{gAnswer2014} &{$0.29$} & {$0.32$}  & {$0.43$} \\
       KGQAN~\cite{Omar-etal-2021-universal} &{$0.49$} & {$0.39$}  & {$0.43$} \\
       SLING~\cite{Nandana-etal-2020-Leveraging} &{$0.39$} & {$0.50$}  & {$0.44$} \\
       NSQA~\cite{kapanipathi-etal-2021-leveraging} &{$0.31$} & {$0.32$}  & {$0.45$} \\
       Zheng et al.~\cite{zheng2019question} &{$0.45$} & {$0.47$}  & {$0.46$} \\
       GenRL~\cite{Rossiello-etal-2021-Generative} &{$0.49$} & {$0.61$}  & {$0.53$} \\
       {Our Approach} & {$\bf0.77$}  & {$\bf0.67$}    & {$\bf0.72$}   \\\hline  %
                    
    \end{tabular}
    \label{tab:compml}  
    \vspace{-5mm}
\end{table}

\paragraph{Comparison with SOTA:}
In Table \ref{tab:compml}, we compare our approach with the most recent QA systems evaluated on QALD-9. TeBaQA~\cite{Vollmers-etal-2021-Knowledge} maps NL questions to SPARQL queries through learning templates from the QALD-9 dataset. However, the training dataset is small, consisting of only 403 questions, which limits the approach's ability to learn templates. gAnswer~\cite{gAnswer2014} and EDGQA~\cite{Xixin-etal-2021-EDG-Based} are graph-based approaches that interpret an NL question into a semantic query graph containing an edge for each relation mentioned in the question. SLING~\cite{Nandana-etal-2020-Leveraging} and GenRL~\cite{Rossiello-etal-2021-Generative} are relationship linking frameworks developed for QALD. These approaches achieve the highest $F_1$ scores (ranging from $0.40$ to $0.55$) among all systems evaluated on the QALD-9 dataset. Our compositional approach outperforms all these methods, achieving an $F_1$ score of $0.72$.

\section{Related Work}
\label{sec:related_work}

Some QALD systems (such as ORAKEL~\cite{CIMIANO2008325}, Pythia~\cite{Unger2011PythiaCM}, QueGG~\cite{Elahi2024,EnglishSemantic2021,DBLP:conf/iswc/Benz2020}, and LexExMachinaQA~\cite{LexExMachinaQA}) use Lemon lexica for lexicalization.
For instance, Pythia~\cite{Unger2011PythiaCM} is built on Lexicalized Tree Adjoining Grammars~\cite{DBLP:conf/tag/UngerHC10} (LTAG) as a syntactic formalism and DUDES~\cite{cimiano-2009-flexible} for specifying semantic representations.  The QueGG system~\cite{EnglishSemantic2021,DBLP:conf/iswc/Benz2020} automatically generates a QA grammar from manually-created Lemon lexica. This grammar is then used to transform questions into SPARQL queries. However, the approach uses manually created sentence templates to cover syntactic variations and has very limited support for complex questions. 
The QueGG system~\cite{Elahi2024} was compared with GPT-3.5 Turbo in a zero-shot scenario by prompting it with an instruction to generate a SPARQL query for a question related to DBpedia, without providing any lexical information. %
None of these approaches systematically evaluated the impact of lexical information on QALD performance. In contrast, our compositional approach uses dependency parsing, requiring no handwritten sentence templates.
It addresses compositionality in a principled way using DUDES in combination with a tree merging and scoring component, covering a wide variety of complex questions. 

WDAqua-core1 system\cite{DBLP:journals/semweb/DiefenbachBSM20} maps  natural language sentences to KB elements by comparing an n-gram with the \texttt{rdfs:label} of an entity. For instance, the approach maps the natural language term \emph{``writer''} to \emph{dbo:writer} but fails to map it to other variations such as \emph{dbo:creator} or \emph{dbo:composer}. Some QALD systems (such as AskNow \cite{AskNow2016}, DEANNA \cite{DEANNA2012}, Sem\-QALD \cite{semanticParsing}, QAKiS \cite{QAKiS_2012}, QAnswer \cite{QAnswer2015} etc.) use pattern dictionaries (e.g., BOA~\cite{BOA} or similar dictionaries) that map natural language terms to KB elements, while other approaches (Xser~\cite{AskNow2016}, gAnswer~\cite{AskNow2016}, CASIA~\cite{AskNow2016}) use relational lexicalizations (e.g., PATTY~\cite{nakashole-etal-2012-patty}). The resources and dictionaries are very limited, and none of these approaches has investigated the impact of lexicalization on QALD.

One major limitation of state-of-the-art QALD systems is the lack of semantic compositionality for dealing with complex queries. To handle complex questions, Wang et al.~\cite{MultiRelationQuestions2024} proposed a model that uses graph convolutional networks (GCNs) ~\cite{GraphConvolutional2024} and performs reasoning over multiple KG triples. Similarly, Shekarpour et al.~\cite{Shekarpour-2012} use a combination of KB concepts with a HMM model. The approach first finds the segment (e.g., \emph{``mayor''}, \emph{``capital''}, \emph{``Russia''}) of a query (e.g., \emph{``Who is the mayor of the capital of Russia?''}) and then maps them to the appropriate resources. Other approaches (such as GETARUNS~\cite{delmonte2008computational}, IBM Watson~\cite{IBM-2012} etc.) generate a logical form from a query. The approach generates triple patterns that are then split up again as properties are referenced by unions, resulting in many combinations of triples and wrong SPARQL queries. In contrast, our compositional approach selects the correct SPARQL query using a SPARQL selector (detailed in Section \ref{sec:SPARQL Selector}) from all possible combinations of SPARQL queries.
There are rule-based architectures~\cite{ForecastTKGQuestions2023,dubey2021towards} to deal with complex questions, but the coverage of these approaches is completely limited to the rules added based on linguistic heuristics and observed patterns in the data.

\section{Conclusion and Future Work}

We have investigated the role and impact of explicitly given lexical knowledge in the context of QALD systems. 
We have presented a novel compositional system that uses this knowledge and demonstrated that it achieves performances in terms of micro $F_1$ scores well beyond the current state-of-the-art. In fact, our approach achieves a micro $F_1$ score of $0.72$, which is $0.19$ higher than the performance of the best state-of-the-art system on QALD ($0.53$). 
In this regard, our work has to be understood as providing a proof of concept that shows the impact of lexical knowledge and of a compositional approach.

Our approach handles complex queries using DUDES for semantic composition, combined with a tree merging and scoring component and a SPARQL selector, thereby covering a wide variety of complex questions. All we need is a lexicon in Lemon format.
At the same time, our results show that LLMs are very limited in their ability to compose, as they cannot leverage provided lexical knowledge to the same extent as our proposed approach. Overall, our results suggest new avenues for QALD research by highlighting the role of explicit lexical knowledge and compositionality. However, there are also limitations.

First, a necessary prerequisite for our approach is the availability of a Lemon lexicon~\cite{lemon}, which is manually created and takes approximately $16$ hours to produce for $599$ lexical entries.
Therefore, future work will focus on automating this process using one of the approaches, such as LexExMachina~\cite{EllEC21} and M-ATOLL~\cite{M-ATOLL-2014}, which automatically create a lexicon for the QA system.
Another limitation is combinatorial explosion, i.e., the exponential growth of combinations when multiple candidate DUDES exist across multiple nodes of a tree, which increases response times considerably. 

We provided a promising direction of QALD for future work consisting of the development of a hybrid system that combines the benefits of a compositional approach with the generalization abilities of large language models to bridge the lexical gap while leaving composition to a symbolic approach.

\subsubsection{Acknowledgements and Funding.}

This work is partially funded by the Ministry of Culture and Science of the State of North Rhine-Westphalia under grant no NW21-059A (SAIL).

This preprint has not undergone peer review (when applicable) or any post-submission improvements or corrections. The Version of Record of this contribution is published in Knowledge Engineering and Knowledge Management,  Lecture Notes in Computer Science (LNCS, volume 15370), and is available online at \url{https://doi.org/10.1007/978-3-031-77792-9_7}.

\bibliographystyle{splncs04}
\bibliography{bibliography}

\end{document}